\documentclass[11pt,a4paper]{article}
\usepackage[hyperref]{eacl2021}
\usepackage{times}
\usepackage{latexsym}

\usepackage{microtype}
\usepackage{graphicx}
\usepackage{multirow}
\usepackage{booktabs}
\usepackage{xcolor}
\usepackage{wrapfig}
\usepackage{eurosym}
\usepackage[english,spanish,swedish,portuges,german]{babel} 
\usepackage{pbox}
\usepackage{subfig}

\usepackage{scalefnt}
\usepackage{xcolor}
\usepackage{color,soul}
\setul{0.5ex}{0.3ex}
\setulcolor{red}

\aclfinalcopy

\title{Emotion Ratings:\\ How Intensity, Annotation Confidence and Agreements
are Entangled}

\author{Enrica Troiano, Sebastian Pad\'o \and Roman Klinger\\
 Institut f\"ur Maschinelle Sprachverarbeitung \\
  University of Stuttgart, Germany \\
  {\tt \{firstname.lastname\}@ims.uni-stuttgart.de}\\
}

\date{}

\begin{document}
\maketitle
\begin{abstract}
  When humans judge the affective content of texts, they also
  implicitly assess the correctness of such judgment, that is, their
  confidence.  We hypothesize that people's (in)confidence that they
  performed well in an annotation task leads to (dis)agreements among
  each other.  If this is true, confidence may serve as a diagnostic
  tool for systematic differences in annotations.  To probe our
  assumption, we conduct a study on a subset of the Corpus of
  Contemporary American English, in which we ask raters to distinguish
  \textit{neutral} sentences from \textit{emotion}-bearing ones, while
  scoring the confidence of their answers.  Confidence turns out to
  approximate inter-annotator disagreements. Further, we find that
  confidence is correlated to emotion intensity: perceiving stronger
  affect in text prompts annotators to more certain classification
  performances.  This insight is relevant for modelling studies of
  intensity, as it opens the question wether automatic regressors or
  classifiers actually predict intensity, or rather human's
  self-perceived confidence.
\end{abstract}
\section{Introduction}
A plethora of theories exist on the matter of emotions: the intensity
of affective states, their link to cognition, and their arrangement
into categories are just a few of the angles from which psychology has
tackled this complex phenomenon
\cite{gendron2009reconstructing}. Correspondingly, in computational
emotion analysis, texts have been associated to values of intensity
\cite{strapparava-mihalcea-2007-semeval,mohammad-bravo-marquez-2017-wassa},
to cognitive components \cite{Hofmann2020}, and discrete classes
\cite[][i.a.]{zhang2018text,zhong-etal-2019-knowledge}. In support of
these tasks, substantial research effort has been directed to resource
construction, which typically relies on the participation of human
judges.  Yet, emotions are a subjective experience. Their
interpretation in text varies across individuals, and this poses a
major challenge in emotion analysis: it is hard to reach acceptable
levels of inter-annotator agreement (IAA)
\cite{bobicev2017inter,Schuff2017,Troiano2019}.

On their side, humans (roughly) know how well they can
\textquotedblleft read\textquotedblright\space emotions
\cite{realo2003mind}.  They judge the affective content of texts, and
at the same time, the correctness of such judgment: in other words,
annotators can evaluate their own confidence with respect to their
labelling decisions.  This hints a possible relation between
confidence and IAA.  One could expect that annotators are more likely
to incur inconsistencies when they feel uncertain about their answers.
Hence, it would be interesting to verify if self-assessed confidence
approximates inter-annotator disagreements in affect-related
annotation tasks. The collection of this type of judgments
is not a common practice: 
past research has found that self-assigned scores of
confidence are predictable based on some vocal attributes of emotion
speech stimuli \cite{lausen2020emotion}, but this has
not been done on text, to the best of our knowledge.

Yet another aspect involved in emotion recognition and which, at first
sight, relates to confidence, is emotion intensity.  It would be
intuitive to assume that emotions are recognized with higher
confidence if they are expressed with stronger magnitude (e.g.,
``\emph{The teacher exploded}'' $>$ ``\emph{He snapped his annoyed
  temper}'', ``\emph{sadder}'' $>$ ``\emph{a bit sad}'',
``\emph{ecstasy}'' $>$ ``\emph{joy}'').  Still, in the sentence
``\textit{We had to cheer him up; later, he was off the ground}'',
readers have to choose what part of the text to attend to (the
challenge that the speakers undertook -- not intense, or the effect
they had -- intense) and be very confident about either
choice. Similar counter examples reveal that the link between the
perception of emotion intensity and the self-perceived confidence is
opaque, and leaves room for exploration.

In this paper, we experimentally investigate the relationship between
three human judgments: about the presence of emotions, about their
intensity, and about the confidence of the annotation decision.
Leveraging such information, we aim at understanding in what cases
annotators differ regarding the judgement that an emotion is
expressed.  Our first research question is: (RQ1) Are (dis)agreements
with respect to the presence of emotions related to confidence?
Second, (RQ2) Are judgments of intensity and confidence entangled?  We
bring these issues together in an annotation study based on emotion
recognition, in which self-perceived confidence is a dimension to be
rated.  Given a subset of the Corpus of Contemporary American English
(COCA) \cite{coca}, raters distinguish emotion-bearing sentences from
neutral ones, while quantifying both the intensity of the emotion and
their confidence on a Likert scale
\cite{begue2019confidence}.
  
We find that confidence can explain systematic differences in
decisions of annotators; so does intensity; and impressions on
intensity and confidence are correlated. Based on these results, we
devise a strategy to leverage the two factors and smoothen out
inter-annotator inconsistencies.

\section{Annotation Setup}

\paragraph{Tasks.} The first step in this study is to collect emotion
assessments.\footnote{The guidelines are in
  the Appendix. Our data is accessible at
  \url{https://www.ims.uni-stuttgart.de/data/emotion-confidence}.}
We are not interested in which emotion people interpret from text, but
rather if they recognize any.  Judges read sentences and answer the
question: \textit{(\textsc{Emo}) Is it Emotional or Neutral?}  For the
items deemed to express an emotion, we also ask \textit{(\textsc{Int})
  How strong is it?}, which enables us to obtain ratings about
affective strengths on a Likert scale from 1 (not intense) to 3
(very). Lastly, since raters interpret emotions without an immediate
first-hand experience, we have them self-evaluate their own judgments
on a scale from 1 (unsure) to 3 (certain), in response to the question
\textit{(\textsc{Conf}) How confident are you about your answer to
  \textsc{Emo}?}

As for \textsc{Emo}, we acknowledge that the emotional content of an
utterance can be inferred from many perspectives. It is possible to
assess one's own emotion after reading the text, to reconstruct the
affective state of the writers who produced it, to guess the reaction
that they intended to elicit in the readers, and so on.  To avoid
confusion, we instruct annotators to consider the presence of an
emotion only with respect to their personal viewpoint.

We opt for an in-lab setting. Raters are three female master students
aged between 24 and 27, who are proficient in English, and have some
annotation experience and background in computational emotion analysis.

\paragraph{Data.}
Corpora that include emotion classes or gradations are tailored on
specific domains, like self-reports \cite{Scherer1997}, tweets
\cite{nrcintensity} and newspapers \cite{strapparava2008learning}.  We
broaden our focus to multiple genres, and annotate sentences from the
2020 version of COCA\footnote{https://www.english-corpora.org/coca/},
which includes unlabelled texts that occurred from 1990 to present in
different domains, like blogs, magazines, newspapers, academic texts,
spoken interactions, fictions, TV, and movie subtitles.

With a corpus of this size ($>$1B words), considering all data points
would be costly, and randomly selecting them may cause imbalance in
the final annotation -- i.e., a majority of \textit{neutral}
instances. Therefore, we draw a sample biased towards emotional
sentences with a combination of rules and classifier-based
information.  To obtain such a classifier, we fine-tune the
pre-trained BERT \citep{Devlin2019} base-case model on a number of
emotion analysis resources\footnote{Details and classifier's
  performance in Appendix.}, adding a classification layer that
outputs the labels \textit{emotion} or \textit{neutral}.  Having that,
we filter academic texts out of COCA for their arguably impartial
language, and from each of the other genres, we randomly pick 500
sentences; out of these, we sample 100 sentences balanced by class,
i.e., 50 labelled as \textit{neutral} by our classifier, 50 as bearing
an \textit{emotion}.  Thus, the annotators are shown 700 items, 100
per domain, with a balanced class distribution, according to the
classifier.

\section{Results}
\begin{table*}
  \centering
  \mbox{}\hfill
  \subfloat[][Cohen's $\kappa$ for
  annotator pairs on \textsc{Emo}.\label{tab:iaa}]{
    \setlength{\tabcolsep}{8pt}
    \begin{tabular}{ll}
      \toprule
      &IAA \\
      \cmidrule{2-2}
      A1--A2&.38\\
      A2--A3&.43\\
      A3--A1&.30\\
      \bottomrule
    \end{tabular}
  } \hfill \subfloat[][Counts of \textit{emotion} (E) and
  \textit{neutral} (N) items for \textsc{Emo} answers, aggregated by
  agreement (1 vs. 2, 3 vs. 0).\label{tab:iaacounts}]{\setlength{\tabcolsep}{8pt}
    \begin{tabular}{r rr}
      \toprule
      & \multicolumn{2}{c}{Counts}\\
      \cmidrule{2-3}
      & 1 vs. 2 & 3 vs. 0\\
      \cmidrule(lr){2-2}\cmidrule(lr){3-3}
      E &138 &304\\
      N &170 &88\\
      \bottomrule
    \end{tabular}
  }
  \hfill
  \subfloat[][Fleiss' $\kappa$ on \textsc{Emo} for each value of
  \textsc{Conf} and \textsc{Int}.\label{fleiss-bins}]{
    \setlength{\tabcolsep}{8pt}
    \begin{tabular}{rrr}
      \toprule  
      & \textsc{Conf}& \textsc{Int} \\
      \cmidrule(r){2-2}\cmidrule(r){3-3}
      1 & $-$.001&.04\\
      2& .03 & .20 \\
      3& .39 & .30 \\
      \bottomrule
    \end{tabular}
  }
  \hfill\mbox{}
  \caption{Inter-Annotator Agreements.}%
  \label{iaa}%
\end{table*}

To answer our research questions, we first observe annotators'
agreements on \textsc{Emo}. The highest \citeauthor{Cohen1960}'s
$\kappa$ \shortcite{Cohen1960} between pairs of human judges was .43
(Table~\ref{tab:iaa}); \citeauthor{fleiss1971measuring}' $\kappa$
\shortcite{fleiss1971measuring} for the three annotators was .34.

At first glance, these numbers appear unsatisfactory. On the one hand,
they are due to the skewed class distribution in the annotators'
choices.\footnote{With skewed class distribution, chance agreement
  increases, penalizing the resulting $\kappa$
  \cite{cicchetti1990high}.} On the other, they can be traced back to
the way in which the \textsc{Emo} task was formulated: asking if a
text is emotional from the readers' own point of view (e.g., it
``describes an event [...] to which \textit{you} would associate an
emotion'', see Appendix) as opposed, for instance, to the writers',
paves the way for more heterogeneous responses.

However, a look at other IAA measures, like the absolute counts of
items that were assigned to each label, leads to a more detailed
picture.  Table~\ref{tab:iaacounts} breaks down the annotated
categories by agreement: column 1 vs. 2 corresponds to the groups of
items on which 1 annotator chose a label, while the majority opted for
the other; column 3 vs. 0 shows how many times all three annotators
agreed. We see that 138 sentences out of 700 were deemed emotionally
charged by only 1 person (and hence, were associated to
\textit{neutral} by the two others).  2 annotators picked the
\textit{emotion} class for 170 sentences, i.e., those which were
\textit{neutral} according to just 1 rater. Overall, as the amount of
considered judgments increases, so does their intersubjective validity
about emotions.  This tendency is clear in column 3 vs. 0, which shows
that there were more emotional instances with 3 identical labels than
those with conflicting ratings.  Indeed, perfect agreement was reached
for 392 items (304 \textit{emotion} and 88 \textit{neutral}), more
than half the data, suggesting that people had a shared understanding.

\subsection{Confidence Approximates Disagreements (and so does Intensity)}
We next focus on the items that received incoherent judgments.
Annotators seem to diverge on the presence/absence of emotions in a
systematic way.  Specifically, their inconsistencies correspond to
certain patterns in the ratings of confidence, as well as intensity.
Taking pairs of annotators, we see that the one who picks the
\textit{emotion} class tends to do that with low
confidence.\footnote{Ratings on disagreements are in Appendix,
  Table~\ref{disagreements}.} As an example representative of the
general trend, on 11 sentences annotator A3 makes the \textit{neutral}
choice, while annotator A1 picks \textit{emotion}, but rates such
items with confidence 1 (5 sentences), or 2 (6 sentences) -- never
using the highest degree of confidence.  The same holds for intensity:
A1 rates 10 of the 11 sentences as having the lowest intensity, and 1
sentence as having intensity 2 -- none with the highest intensity
value.

Hence, to answer RQ1, the evaluation of intensity and the
self-evaluation of confidence underlie disagreements in discrete
emotion annotations. Further, they show that different intuitions are
not totally incompatible, since the annotator who takes the
\textit{emotion} decision does so without being extremely confident,
and gauging intensity as rather weak.

We corroborate this finding by looking at a more standard measure of
IAA, namely \citeauthor{fleiss1971measuring}' $\kappa$, which turns
out to be affected by both \textsc{Conf} and \textsc{Int}.  We compute
it once more for the answers of \textsc{Emo}, but here we consider to
be emotional only the items on which all annotators choose a certain
level of confidence or intensity.  Table~\ref{fleiss-bins} displays
$\kappa$ separately for different levels (rows) of \textsc{Conf} and
\textsc{Int}.  Low values aside, these results inform us that the
lower \textsc{Conf}/\textsc{Int}, the more prominent are
disagreements.  The highest IAA (.39), instead, is achieved for the
most confident answers.

\paragraph{Post-Processing Disagreements.}
\begin{table}
\centering
\addtolength{\tabcolsep}{-2pt}
\begin{tabular}{r rr lll rr}
  \toprule
  &\multicolumn{2}{c}{\textsc{Emo} \textsc{Conf}$<$2}&&&&\multicolumn{2}{c}{\textsc{Emo} \textsc{Conf}$<$3}\\
  \cmidrule{2-3} \cmidrule{7-8}
  & 1 vs. 2 & 3 vs. 0 &&&& 1 vs. 2 & 3 vs. 0\\
  \cmidrule(lr){2-2}\cmidrule(lr){3-3}\cmidrule(lr){7-7}\cmidrule(lr){8-8}
   E &172 &187 &&&& 141&56 \\
  N &169  &172 &&&& 73&430\\
  \bottomrule
  \addlinespace[.3cm]
  \toprule
    &\multicolumn{2}{c}{\textsc{Emo} \textsc{Int}$<$2}&&&&\multicolumn{2}{c}{\textsc{Emo} \textsc{Int}$<$3}\\
  \cmidrule{2-3} \cmidrule{7-8}
  & 1 vs. 2 & 3 vs. 0 &&&& 1 vs. 2 & 3 vs. 0\\
  \cmidrule(lr){2-2}\cmidrule(lr){3-3}\cmidrule(lr){7-7}\cmidrule(lr){8-8}
   E &165 &82 &&&& 57&7 \\
  N & 118&335&&&& 17&619\\
  \bottomrule
\end{tabular}
\caption{Counts of labels
for subsets of ratings on \textsc{Emo}, post-processed with 
acceptance thresholds $<$2 and $<$3,
for both \textsc{Conf} (top) and \textsc{Int} (bottom).}
\label{absolute-numbers}
\end{table}
If systematic differences among annotators can be diagnosed with the
help of confidence and intensity, can they also be resolved to some
extent?  We use the \textsc{Conf} and \textsc{Int} scores as
acceptance thresholds for the label \textit{emotion}, so to
post-process the \textsc{Emo} decisions of each judge: they turn into
\textit{neutral} if the corresponding \textsc{Conf} or \textsc{Int}
answer does not reach a certain threshold t.  For instance, using
\textsc{Int} as threshold, with t$<$2 all items labeled
\textit{emotion} in \textsc{Emo} are kept as such only in case the
\textsc{Int} is 2 or more, all the others are mapped to
\textit{neutral}.

Agreement counts on the post-processed annotation of \textsc{Emo} are
in Table~\ref{absolute-numbers}.  We see, again, that the number of
agreed upon items increases by increasing the sets of equal ratings.
For instance, 283 sentences received 2 unanimous judgments (column 1 vs. 2,
under \textsc{Emo} \textsc{Int}$<$2), and 417 received 3.
In comparison to the original annotation in Table~\ref{tab:iaacounts},
we can observe a considerable change in the number of items with
perfect agreement.  While in the raw judgments they were 392, with
t$<$3 they increase to 626.  We find a similar pattern when leveraging
confidence: with t$<$2 (low confidence), it is obtained for 359 items,
and with t$<$3 (moderate), perfect agreement increases to 486
items.

This comes at the cost of agreeing on fewer \textit{emotion} sentences (304
before filtering, 7 and 56 after applying the highest threshold to \textsc{Int} and
\textsc{Conf}), but it indicates that the better raters agree on intensity or 
confidence, the more they agree regarding the presence or absence of 
emotions.

\subsection{Stronger Intensity, Higher Confidence}
Having found that confidence and intensity have a similar relationship
to disagreements, we move to analyzing how they link to one
another. To address RQ2, we focus on the ratings of the 304 sentences
with the unanimous \textit{emotion} judgment.  For them, we compute
the intra-annotator correlation between the answers to \textsc{Int}
and the corresponding ratings of \textsc{Conf}.  A Spearman's $\rho$
\citep{spearman1961proof} of .5 for annotator A1, .58 for A2 and .64
for A3 (p-value $<$.05 for all) reveals a significant positive
correlation between intensity and confidence. This suggests that
people believe they correctly classified a text if they also perceived
high emotion intensity.

\begin{figure}
  \centering
  \includegraphics[width=\linewidth]{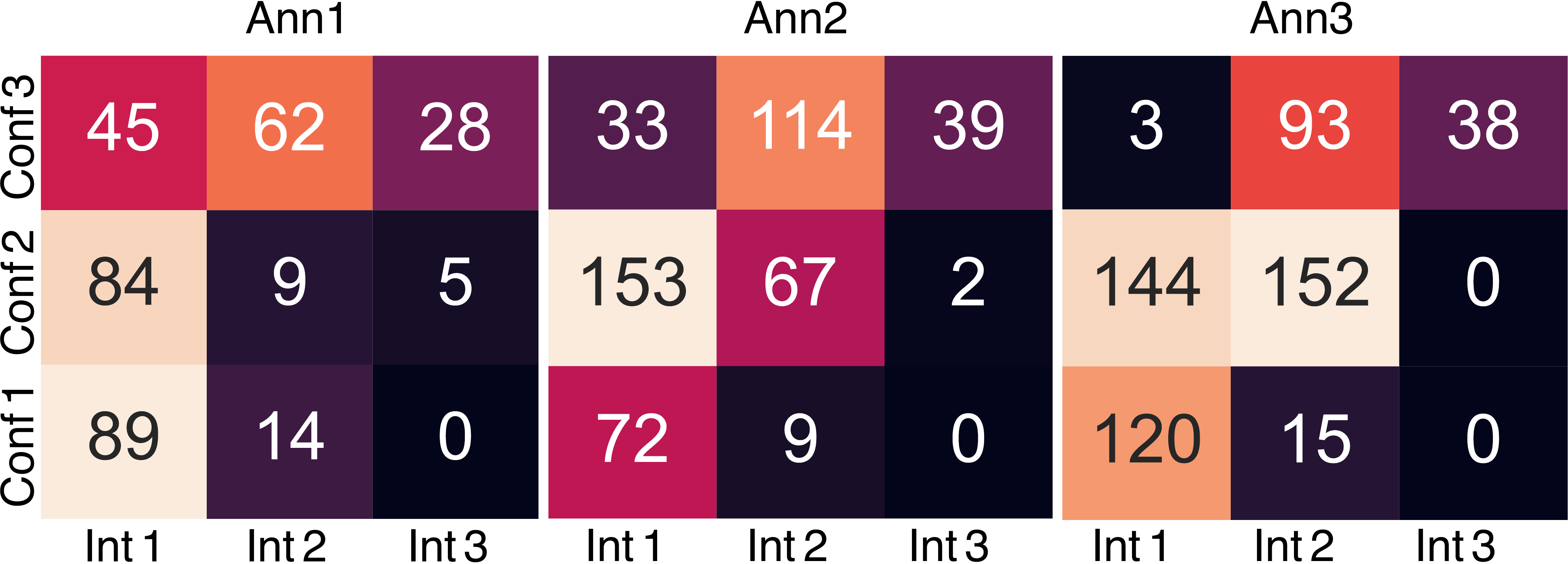}
  \caption{Cross-tabulation of \textsc{Int} and \textsc{Conf} by annotator,
    for the items that each of them deems emotional.}
  \label{frequencies}
\end{figure}

Figure~\ref{frequencies} gives an in-depth account of the
\textsc{Conf}-\textsc{Int} relation.  It plots the counts of items
that were labeled with a certain emotion intensity together with a
certain confidence level, separately for each annotator. The columns
\textsc{Int3} tell us that rarely annotators perceive intensity as
strong without being extremely confident that the text
expresses an emotion. In fact, no instance was rated with the highest
intensity and the lowest confidence (\textsc{Int}3-\textsc{Conf}1) at
the same time. Conversely, for cases of low intensity, annotators tend
to stay low also on the scale of confidence.

\paragraph{On what do People Agree?}
On the 304 sentences considered emotional by all, if the annotators
gave the same score to intensity (i.e., perfect agreement is both on
\textsc{Emo} and \textsc{Int}), they did not have a total disagreement
on confidence (i.e., there are always at least 2 people with the same
\textsc{Conf}) and vice versa.  This may be a sign of the correlation
between the two variables.
 
Moreover, some items were scored with perfect agreement on all
questions. It is the case of sentences like \textquotedblleft
\textit{I can't believe that you saved my life}\textquotedblright,
considered, with the highest level of confidence, to convey an emotion
of intensity 3, \textquotedblleft \textit{Get off my
  back!}\textquotedblright\space deemed to have a mild intensity, and
\textquotedblleft \textit{,,You have such an interesting life,"\space
  she said, after a little small talk}\textquotedblleft\space with
intensity 1.  While these examples are rated as highly certain
(\textsc{Conf}3), there are also sentences on which people agree
across all confidence-intensity combinations.\footnote{A detailed
  analysis is in Table~\ref{example-sentences}, Appendix.}  None of
the instances has \textsc{Conf}1 and \textsc{Int}3, while a number of
examples have \textsc{Conf}3 and \textsc{Int}1, indicating that it is
harder to be uncertain of a strong emotion than to be sure of a weak
one.

\section{Discussion and Conclusion}

This annotation investigated if the perceived emotion (class), a
perceived feature of emotion (intensity), and self-perception
(confidence) are tied together -- and can help understand inconsistent
annotations. We found that (RQ1) both intensity and confidence account
for inter-annotator inconsistencies relative to binary decisions.
Adopting confidence as an acceptance threshold showed that higher
scores lead to more uniform assessments of emotions; though not a
surprising effect of confidence, this also applies to intensity.
Moreover, (RQ2) the two variables are correlated, that is, people feel
more certain about their emotion recognition performance on items with
high intensity.

We acknowledge that our design of \textsc{Emo} naturally incurs the
risk of inconsistent answers. However, precisely for the subjective
nature of the task, the finding that disagreements decrease with high
\textsc{Conf}/\textsc{Int} is interesting in itself: some judgments
which are seemingly unsolvable can be explained by certain perceived
properties of emotions (intensity) or self-perceived features
(confidence).

From these results we can draw some lessons.  First, the correlation
between confidence and intensity brings relevant implications to all
those studies that focus on emotional strength.  When asked to
evaluate intensity, do people confound that with confidence?  Even
more, is there a causal relation between the two?  As a best practice
to put safeguards in their guidelines, experimenters may ask people to
tease the two variables apart. Potentially, this issue concerns
modelling studies as well: do classifiers for emotion intensity
predict such feature, or rather the confidence with which people judge
emotions?  We provided reasons to look into this further.

Second, confidence turned out to be an important dimension of rating,
because it can inform us when the annotators expect to disagree.  When
judgments diverge, annotators do not deem their intuition credible.
Hence, our finding that confidence approximates disagreements means
precisely that people themselves predict their performance to differ
from that of the others.

Concretely, all this knowledge can come in support of annotation
studies.  Including confidence as a rating dimension may give an
additional source of information about annotators' reliability. This
can help experimenters to refine the guidelines in a pre-testing
phase: one might want to disentangle cases in which annotators'
disagreements are random -- signaling a lack of annotators'
reliability, and when, instead, they are due to consistently different
ways of perceiving and reporting on confidence.  In this second case,
disagreements may be normalized to an extent, by post-processing the
annotation results.  For instance, as people seem to agree on the
class \textit{emotion} only if they also agree on certain degrees of
\textsc{Conf}, there might be some levels of confidence (or intensity)
that one filters in/out of the final annotation labels.

A similar strategy may be somewhat restrictive, since it accepts as
emotional only those items on which humans' intuition fall above a
pre-defined threshold.  While we observed agreement on such items, it
is possible to adopt more nuanced evaluation approaches and integrate
information about intensity or confidence into IAA measures. As an
example, disagreements between two raters can be penalized more when
the one choosing the \textit{emotion} label does so by perceiving
extreme confidence or intensity -- even though we provided evidence
that these cases are rare. Future work could explore this direction.

In summary, we uphold that disagreements are not necessarily
symptomatic of unreliability. This claim has so far not found much
attention in emotion annotations, but is in line with a more general
body of research dedicated to the reasons and the patterns underlying
annotators' disagreements and to the ways in which their intuitions
should be aggregated and evaluated
\cite[i.a.]{bayerl-paul-2011-determines,bhardwaj-etal-2010-anveshan,qing-etal-2014-empirical,peldszus-stede-2013-ranking,plank-etal-2014-linguistically,sommerauer-etal-2020-describe}. The
applicability of these ideas to emotions should not come as a surprise
--- their assessment can derive from perceptive and meta-perceptual phenomena
(intensity and confidence, for instance).  Therefore, if emotion
judgments alone might not be sufficient to measure the quality of
annotations, they can be enriched and, eventually, explained by the
knowledge of such phenomena.

\section*{Acknowledgements}
This work was supported by Deutsche Forschungsgemeinschaft (project
CEAT, KL 2869/1-2) and the Leibniz WissenschaftsCampus T\"ubingen
``Cognitive Interfaces''.  We thank Laura Oberl\"ander, Agnieszka
Faleńska and the anonymous reviewers for their constructive feedback.

\clearpage

\bibliographystyle{acl_natbib}
\bibliography{lit}

\appendix

\clearpage
\section{Annotation Guidelines}
In what follows are the detailed guidelines of the annotation. Q1
corresponds to \textsc{Emo} in the paper, Q2 corresponds to
\textsc{Conf}, and Q3 to \textsc{Int}.

In Q1, annotators were required to disregard the emotion that the
author of the utterances intended to express or to elicit in
others. Their task was to give their immediate, personal impression
with respect to the presence of an emotion.

Before annotating the 700 sentences 
selected from COCA, we observe inter-annotator
agreement on a pre-annotation trial. 
With 70 sentences, Cohen's $\kappa$ for pairs of annotators
was found to be satisfactory (.52, .6 and .43) and
motivated us to complete the job on the
rest of the sentences. The job was completed
upon a compensation of $60$\euro.

\subsection{Task Description\\ Approx Duration: 3 hours}
In this annotation trial, you will assess if texts are emotional or
neutral.\\\\\textbf{Neutral} sentences are those
which \begin{enumerate}
	\item bear no affective connotation.\\
\end{enumerate}
\textbf{Emotional} sentences are those which either
\begin{enumerate}
  \setcounter{enumi}{1}
\item describe an event, a concept or state of affairs to which you would associate an emotion;
\item have an emotion as a central component of their meaning.\\
\end{enumerate}
Examples of $1.$ are: \\
\begin{minipage}{.48\textwidth}

 \hfill \textit{I am wearing my mask.}

\hfill \textit{She answered her phone.}

\hfill \textit{A new deal was established between the parties.}

\hfill \textit{The elections are over.}\\
\end{minipage}
Examples of $2.$ are: \\
\begin{minipage}{.48\textwidth}

\hfill \textit{I saw my bestfriend.}

\hfill \textit{She was being pretty arrogant to me.}

\hfill \textit{A war started in Westeros.}

\hfill \textit{The king found an old sausage under his bed.}\\

\end{minipage}
Examples of $3.$ are: \\
\begin{minipage}{.48\textwidth}

\hfill \textit{I am so happy to see you.}

\hfill \textit{She was bursting with arrogance.}

\hfill \textit{And there she was, desperate for her family.}

\hfill \textit{I couldn't stand the catering food, bleark!}\\

\end{minipage}

\subsection{Guidelines}
You will be shown individual sentences and, for each of them, you will
answer $3$ questions (\textbf{Q1}, \textbf{Q2}, \textbf{Q3}). Go for
your immediate reaction to the text -- avoid over-assessments.
\paragraph{Q1} Given a sentence, please ask yourself: \textbf{ is it emotional (E) or neutral (N)?} Type:
\begin{itemize}
\item \textbf{N}, if the text does not convey any emotion, like in Examples of $1.$;
\item \textbf{E}, if an emotion could be inferred from the text, like
  in Examples of $2.$, or an emotion is a central part of the text,
  like in Examples of $3.$
\end{itemize}

\paragraph{Q2} Ask yourself: \textbf{how confident am I about my
  answer to Q1?} Give yourself a rating on a scale from $1$ to
$3$. Indicate if you are
\begin{itemize}
\item \textbf{1}, not confident at all;
\item \textbf{2}, confident;
\item \textbf{3}, sure.
\end{itemize}

\paragraph{Q3} This question only applies if you answered $2$ or $3$
in \textbf{Q1}: in case the sentence expresses an emotion, \textbf{how
  strong is such emotion?} Assess the degree of its intensity on a
scale from $1$ to $3$, where
\begin{itemize}
\item \textbf{1} is mild;
\item \textbf{2} is intense;
\item \textbf{3} is very intense.
\end{itemize}

\begin{table}[b]
  \centering
  \begin{tabular}{llll}\\
    \toprule  
    & P& R& F1 \\
    \cmidrule(r){2-2}\cmidrule(r){3-3}\cmidrule(r){4-4}
    Emotion & .88&.84& .86\\
    Neutral & .89 & .92 & .90 \\
    \bottomrule
  \end{tabular}
  \caption{Binary classification on UED test set.}
  \label{classification-results}
\end{table}

\section{Binary Classification}
After data pre-processing step (i.e. exclusion of academic texts and sentences
containing words that are masked for copyright reasons),
we use a binary classifier as a guide for the selection of sentences from COCA. 
Using HugginFace\footnote{https://huggingface.co/transformers/},
we fine-tune the pre-trained BERT base-case model on data for emotion recognition,
adding a classification layer that outputs the labels
\textit{emotion} or \textit{neutral}. 
Data are the resources by \newcite{Liu17}, \newcite{Troiano2019},
\newcite{Scherer1997}, \newcite{Ovesdotter2005},
\newcite{li2017dailydialog}, \newcite{ghazi2016detecting},
\newcite{mohammad-bravo-marquez-2017-wassa},
\cite{mohammad-2012-emotional} and \newcite{Schuff2017}.  We make
their format homogeneous with the tool made available by
\newcite{Bostan2018}; next, as the labels in the resulting unified
emotion data (UED) are not binary, we map \textit{neutral} and
\textit{no emotion} instances into \textit{neutral}, and the rest into
\textit{emotion}.  The total 136891 sentences are then split into
train (70\%), validation (10\%) and test (20\%) sets.  Classifier's
performance is in Table~\ref{classification-results}.

\section{Further Analysis}
\subsection{Disagreements}

\begin{table*}
\centering  
\addtolength{\tabcolsep}{-2pt}
\begin{tabular}{l lll lll l lll lll l lll l lll lll}
  \toprule
  &\multicolumn{6}{c}{A1} &&\multicolumn{6}{c}{A2} && \multicolumn{6}{c}{A3}\\
  \cmidrule{2-7}\cmidrule{9-14}\cmidrule{16-21}
  &\multicolumn{3}{c}{\textsc{Conf}}&\multicolumn{3}{c}{\textsc{Int}}&&\multicolumn{3}{c}{\textsc{Conf}}&\multicolumn{3}{c}{\textsc{Int}}&&\multicolumn{3}{c}{\textsc{Conf}}&\multicolumn{3}{c}{\textsc{Int}}\\
  \cmidrule(r){2-4}\cmidrule(r){5-7}\cmidrule(r){9-11}\cmidrule(r){12-14}\cmidrule(r){16-18}\cmidrule(r){19-21}
  & 1 & 2 & 3 &1&2&3&& 1 &2 & 3&1&2&3&& 1 &2 & 3&1&2&3\\
  \cmidrule(r){2-2}\cmidrule(r){3-3}\cmidrule(r){4-4}\cmidrule(r){5-5}\cmidrule(r){6-6}\cmidrule(r){7-7}\cmidrule(r){9-9}\cmidrule(r){10-10}\cmidrule(r){11-11}\cmidrule(r){12-12}\cmidrule(r){13-13}\cmidrule(r){14-14}\cmidrule(r){16-16}\cmidrule(r){17-17}\cmidrule(r){18-18}\cmidrule(r){19-19}\cmidrule(r){20-20}\cmidrule(r){21-21}
  A1-A2& 15  & 8 & 1 &23&1& 0 &&54&  95&28& 149&27&1& &--& -- &--& --&--&--\\
  A2-A3& --& -- &--& --&--&--&& 13 & 25& 6&37&7&0&& 56 & 61& 3&94&26& 0\\
  A3-A1&5& 6&0 &10&1&0&& --& -- &--& --&--&--&& 95 & 128 & 17  &169&70&1\\
  \bottomrule
\end{tabular}
\caption{Distribution of \textsc{Int} and \textsc{Conf} for
  disagreements in the \textsc{Emo} task. For a pair of annotators
  (rows), disagreements are counted when either annotators (i.e., on
  the columns) chooses the \textit{emotion} class.}
\label{disagreements}
\end{table*}

Table~\ref{disagreements} reports the distribution of the scores of
confidence and intensity for the items where the annotators
disagree. This is observed on annotator pairs.  A row considers all
those items on which either annotators (on the columns) chooses the
\textit{emotion} label and the other selects the \textit{neutral} one.

For instance, A1--A2 disagree in total 201 times: on 24 sentences,
A1 makes the \textit{emotion} choice, and on 177 sentences it is
A2 who picks the class \textit{emotion}.  Out of the 24 items, A1
rated 23 as having low intensity and 1 as medium intensity; out of the
177 sentences, 149 are considered of low intensity by A2, 27 as
mild, and only 1 as highly intense.

What emerges overall is that people rarely disagree when the
\textit{emotion}-leaning annotator has extreme confidence, or
perceives very high intensity.

\subsection{Agreements}

\begin{table*}
  \centering
  \addtolength{\tabcolsep}{-4pt}    
  \begin{tabular}{l ll l}
    \toprule
    Text Genre & \textsc{Conf} & \textsc{Int}&Text\\
	 \cmidrule(l){1-1}\cmidrule(l){2-2}\cmidrule(l){3-3}\cmidrule(l){4-4}
    Magazine &1&1& \pbox{20cm}{I was always a little wary of Arya and
                   Sansa  (who also did a little\\
    Stoneheart-style vengeance last year) taking on their mother s role .}\\
    News&1&1&You can't stress because you just have no idea what 's going to happen.\\
    TV&1&1&So maybe Willy's hanging around the wawa one night , looking for a party...Yeah .\\\\
    
    TV &1&2&Mmm , Lordy , Lordy , Lord have mercy .\\
    Web&1&2&\pbox{20cm}{Frost is trying to reconcile impulse with a conscience that needs goals and harbors\\deep regrets.}\\
    Magazine&1&2&\pbox{20cm}{Nature always solves her own problems ; and we can go far toward solving\\our own if we will listen to her teachings and consort with those who love her.}\\\\
    
    Fiction&2&1&\textquotedblleft I'm fine ,\textquotedblright\space he replies absently , eyeing the open book .\\
    Web & 2&1& My sister likes her map : ) HI Chery : ) lol I'll take'em where I can get'em ...\\
    Magazine&2&1&\pbox{20cm}{Whereas coping well means dealing successfully with problems and setbacks,\\savoring-glorying in what goes right-is an equally crucial emotional competence.}\\\\
    
    Blogs&2&2&The soldier talks about child detainees .\\
    TV&2&2&We did n't get to bury the others .\\\\
    
    TV &3&1&I bruised my lip .\\
    Fiction&3&1& \textquotedblleft You have such an interesting life,\textquotedblright\space she said, after a little small talk\\
    Magazine&3&1& \textquotedblleft Chalk is unforgiving, \textquotedblright\space says Oates .\\\\
    
    News&3&2 &They looked happy, confident .\\
    Blogs&3&2 & I am constantly traveling for my job with DISH , and I hate missing all my shows .\\
    Blogs&3&2 &I hope I can work through my feelings and keep his friendship in my life.\\\\
    
    Web&3&3 &\pbox{20cm}{I will completely destroy them and make them an object of horror and scorn ,\\ and an everlasting ruin.}\\
    Spoken&3&3 &\textquotedblleft We 're very worried .\textquotedblright \\
    Spoken&3&3 &\pbox{20cm}{If -- if I could die and bring her back , I would , but I can't , and I have to deal\\ with that now.}\\ 
    
    \bottomrule
  \end{tabular}
  \caption{Sentences on which the annotators
    reached perfect agreement on \textsc{Emo}, \textsc{Conf}, and
    \textsc{Int}.}
  \label{example-sentences}
\end{table*}

A manual analysis of the annotations reveals that perfect agreement
occurs in the presence of certain patterns. Items unanimously
considered emotional often report personal impressions about state of
affairs or the speakers' interlocutors (e.g.,
\textquotedblleft\textit{Paris is so sexy}\textquotedblright,
\textquotedblleft\textit{Your expression changed from excited puppy to
  crestfallen}\textquotedblright), and mostly involve first-hand
experiences of the speakers themselves (e.g.,
\textquotedblleft\textit{We'll miss you, but we'll be
  watching}\textquotedblright, \textquotedblleft\textit{I'm afraid I
  don't see anything very beautiful right now}\textquotedblright,
\textquotedblleft\textit{Others helped me and it made a huge
  difference}\textquotedblright ). Instead, sentences that received 3
\textit{neutral} labels seem to be centered on factual statements,
like \textquotedblleft\textit{Furthermore, the types of materials of
  manufacture are different}\textquotedblright,
\textquotedblleft\textit{They continue walking}\textquotedblright.

One difference between the \textit{emotion} and \textit{neutral}
labels is the frequency of agreement, as we found that people concur
more on the former -- and this invalidates our expectation that, not
being given a varied set of affective categories, and not identifying
\textit{what} emotion they are judging, people would tend to resort to
the neutral choice.  Moreover, annotators converge more on one or more
on the other label depending on the genre of a text: looking at the
distribution of the 304 unanimous \textit{emotion}s (magazine: 28
sentences, blogs: 44, news: 27, tv: 67, fiction: 54, spoken: 39, web:
45) and the 88 \textit{neutral}s (magazine: 18 sentences, blogs: 12,
news: 22, tv: 4, fiction: 5, spoken: 12, web: 15), we see that people
recognize that affect often manifests itself in fictions, for
instance, but is rarer in news -- the opposite holds for the neutral
expressions.

An obvious strategy to recognize emotions would be to find an emotion
name in text. But this is not the case. Sentences that contain emotion
words considered less emotionally intense than others: the majority of
sentences with \textsc{Conf}3-\textsc{Int}2 contain emotion words
(e.g.,\textquotedblleft\textit{I was sad to leave.}\textquotedblright,
while those with \textsc{Conf}3-\textsc{Int}3 are related to extremely
negative states of mind (e.g.,\textquotedblleft\textit{[...] if I
  could die and bring her back , I would , but I can't , and I have to
  deal with that now}\textquotedblright).

In Table~\ref{example-sentences}, we report some example sentences on
which annotators reached perfect agreement across all
confidence-intensity combinations, having choosen the label
\textit{emotion}.  The sentences are extracted from a number of genres
in COCA, and are associated to different scores of intensity
(\textsc{Int}) and confidence (\textsc{Conf}).  Note that there is no
instance that elicited a high intensity evaluation (3) and a low
confidence (either 2 or 3) in the annotators. Instances of
\textsc{Conf}3 and \textsc{Int}1 show that the correlation between
confidence and intensity, though intuitive, has counterexamples.

\end{document}